\documentclass[10pt,journal,letterpaper,compsoc,twocolumn]{IEEEtran}

\usepackage{cite}
\usepackage{graphicx}
\usepackage{psfrag}
\usepackage{subfigure}
\usepackage{url}
\usepackage{stfloats}
\usepackage{amssymb,amsmath}
\usepackage{amsfonts}
\usepackage{array}
\usepackage{CJK}
\usepackage{setspace}
\usepackage[ruled]{algorithm2e}
\usepackage{cases}
\usepackage{listings}

\begin{document}

\title{A New Approach to Speeding Up Topic Modeling}

\author{Jia~Zeng,~\IEEEmembership{Senior Member,~IEEE},
Zhi-Qiang~Liu
and Xiao-Qin Cao
\IEEEcompsocitemizethanks{\IEEEcompsocthanksitem
J.~Zeng is with the School of Computer Science and Technology,
Soochow University, Suzhou 215006, China.
To whom correspondence should be addressed.
E-mail: j.zeng@ieee.org.
\IEEEcompsocthanksitem
Z.-Q.~Liu and X.-Q.~Cao are with the School of Creative Media,
City University of Hong Kong, Hong Kong, China.
}
}

\IEEEcompsoctitleabstractindextext{

\begin{abstract}
Latent Dirichlet allocation (LDA) is a widely-used probabilistic topic modeling paradigm,
and recently finds many applications in computer vision and computational biology.
In this paper,
we propose a fast and accurate batch algorithm,
active belief propagation (ABP),
for training LDA.
Usually batch LDA algorithms require repeated scanning of the entire corpus and searching the complete topic space.
To process massive corpora having a large number of topics,
the training iteration of batch LDA algorithms is often inefficient and time-consuming.
To accelerate the training speed,
ABP actively scans the subset of corpus and searches the subset of topic space for topic modeling,
therefore saves enormous training time in each iteration.
To ensure accuracy,
ABP selects only those documents and topics that contribute to the largest residuals within the residual belief propagation (RBP) framework.
On four real-world corpora,
ABP performs around $10$ to $100$ times faster than state-of-the-art batch LDA algorithms with a comparable topic modeling accuracy.
\end{abstract}

\begin{IEEEkeywords}
Latent Dirichlet allocation, dynamic scheduling, lossy learning, topic models, residual belief propagation, active belief propagation,
message passing, hypergraph, hypergraph, Gibbs sampling, variational Bayes.
\end{IEEEkeywords}

}

\maketitle

\IEEEdisplaynotcompsoctitleabstractindextext

%
\IEEEpeerreviewmaketitle

\section{Introduction} \label{s1}

The probabilistic topic modeling techniques divide the non-zero elements in document-word matrix into several thematic groups called topics,
which have have been applied widely in text mining, computer vision, and computational biology~\cite{Blei:12}.
As the simplest topic model,
latent Dirichlet allocation (LDA)~\cite{Blei:03} represents each document as a mixture of topics
and each topic as a multinomial distribution over a fixed vocabulary.
From the labeling point of view,
LDA assigns the hidden topic labels to explain the observed words in document-word matrix~\cite{Zeng:11}.
This labeling process defines a joint probability distribution over the hidden labels and the observed words.
Employing the Bayes' rule,
we can infer topic labels from observed words by computing
the posterior distribution of the hidden variables given the observed variables from their joint probability.
Such inference techniques have been widely used for learning probabilistic graphical models within the Bayesian framework~\cite{Bishop:book}.
Recent approximate inference methods for LDA~\cite{Asuncion:09} fall broadly into three categories:
variational Bayes (VB)~\cite{Blei:03},
collapsed Gibbs sampling (GS)~\cite{Griffiths:04},
and loopy belief propagation (BP)~\cite{Zeng:11}.

However,
all above batch LDA inference algorithms require multiple iterations of scanning the entire corpus and searching the complete topic space.
So,
the computational cost increases linearly with the number of documents $D$,
the number of topics $K$,
and the number of training iterations $T$.
For massive corpora containing the large number of topics,
these batch LDA algorithms are often inefficient for fast topic modeling because of the high per-iteration cost.
For example,
when $D=10^6$, $K=10^3$ and $T=500$,
GS consumes around forty hours to infer the hidden topics.
Therefore,
we need faster batch LDA algorithms for real-world massive corpora.

In this paper,
we propose a fast batch LDA algorithm,
active belief propagation (ABP).
The basic idea is to actively select a subset of corpus and search a subset of topic space at each training iteration.
ABP is based on the residual BP (RBP)~\cite{Elidan:06} framework that uses an informed scheduling for asynchronous message passing.
First,
we define the message residual as the absolute difference between messages at successive training iterations,
and the residuals are used to evaluate the convergence speed of messages.
The larger residuals the faster convergence speed.
Second,
through sorting residuals at each training iteration,
we actively select those documents and topics that contribute the largest residuals for message updating and passing.
For example,
if ABP selects $10\%$ documents and $10\%$ topics for message passing at each iteration,
it will reach as high as $100$ times faster than BP.
Extensive experiments on four real-world corpora have demonstrated that ABP can achieve a significant speedup
with a comparable accuracy as that achieved by current state-of-the-art batch LDA algorithms.

The proposed ABP algorithm can be interpreted as a fast expectation-maximization (EM) algorithm~\cite{Dempster:77}.
At the E-step,
ABP infers and updates the subset of messages with the largest residuals.
At the M-step,
ABP maximizes the likelihood by estimating parameters based on the subset of updated messages.
After multiple iterations,
it often converges at a local maximum of the joint probability.
Because the residuals reflect the convergence speed of messages,
unlike lazy EM~\cite{Thiesson:01},
ABP focuses all computational resources on inferring and updating those fast-converging messages at each iteration.
Indeed,
the document message residuals approximately follow Zipf's law~\cite{Zipf:book},
where $20\%$ documents contribute almost $80\%$ residuals.
This phenomenon informs us that scanning only a small subset of documents at each iteration by ABP can ensure almost the same topic modeling accuracy as BP.
Since BP can be applied to training other LDA-based topic models~\cite{Zeng:11},
we believe that the basic idea of ABP can be also extended to speed up more LDA-based topic modeling algorithms.

In Section~\ref{s2},
we review some recent fast topic modeling techniques.
In Section~\ref{s3},
we introduce the BP and the residual BP (RBP) algorithms from a novel hypergraph perspective.
Section~\ref{s4} proposes the active belief propagation (ABP) algorithm within the RBP framework.
Section~\ref{s5} compares ABP with the state-of-the-art LDA algorithms on four real-world text corpora.
Finally,
we draw conclusions and envision future work in Section~\ref{s6}.

\section{Fast Topic Modeling Techniques} \label{s2}

Current fast batch LDA algorithms can handle the large number of topics $K$.
For example,
fast GS (FGS)~\cite{Porteous:08} introduces the upper bound on the normalization factor using the H\"{o}lder's inequality.
Without visiting all possible $K$ topics,
FGS is able to draw the target topic sample from the approximate posterior probability normalized by the upper bound.
By refining the upper bound to approximate the true normalization factor,
FGS samples the topic label equal to that drawn from the true posterior probability as GS does.
In this sense,
FGS yields exactly the same result as GS but with much less computation.
Due to the sparseness of the $K$-tuple messages,
sparse GS (SGS)~\cite{Yao:09} partitions the $K$-tuple messages into three parts from which a random topic label can be efficiently sampled,
which avoids additional computations in GS during the sampling process.
GS-based approaches can scale sublinearly in the number of topics by exploiting the fact that
most words never appear in most topics.

Online LDA algorithms~\cite{Banerjee:07,AlSumait:08,Yao:09,Canini:09,Hoffman:10,Wahabzada:11,Mimno:12}
partition the entire corpus into a set of mini-batches.
They infer topic distributions for each unseen mini-batch based on previous topic distributions,
and thus converge significantly faster than batch algorithms,
saving enormous training time by reducing the total number of iterations $T$.
In addition,
online LDA algorithms discard each mini-batch after one look,
and thus require a constant memory usage.
For example,
incremental LDA (ILDA)~\cite{Canini:09} periodically re-samples the topic assignments for previously analyzed words,
or uses particle filtering instead of GS to assign the current topic based on previous topic configurations.
Despite of faster speed,
ILDA does not perform as well as GS.
Online GS (OGS)~\cite{Yao:09} extends SGS to infer topics from unseen data streams.
Online VB (OVB)~\cite{Hoffman:10} combines VB with the online stochastic optimization framework~\cite{Bottou:book},
and can converge to a local optimum of the VB objective function.
However,
OVB requires manually tuning several parameters including the mini-batch size
that determines the best topic modeling performance based on heuristics.
Residual VB (RVB)~\cite{Wahabzada:11} dynamically schedules mini-batches to speed up convergence of OVB.
Sampled online inference (SOI) combines SGS with the scalability of online stochastic inference~\cite{Mimno:12}.
However,
both RVB and SOI cannot process data streams because they re-use previously seen mini-batches.

Parallel LDA algorithms~\cite{Newman:09,Wang:09,Yan:09,Smola:10,Liu:11,Zhai:12}
distribute massive data sets across multiple cores/processors to accelerate topic modeling.
For example,
parallel GS (PGS)~\cite{Newman:09} approximates the asynchronous GS sampling process by
the synchronous updating of the global topic distributions.
Similarly,
FGS and SGS have been also implemented in the PGS parallel architecture.
To reduce communication and synchronization costs in PGS~\cite{Newman:09,Yao:09},
a blackboard architecture has been proposed to deal with state synchronization for large clusters of workstations~\cite{Smola:10}.
Parallel VB (PVB)~\cite{Mahout:10,Zhai:12} uses the Map-Reduce paradigm to speed up LDA.
Although these parallel algorithms can have almost the same topic modeling performance as batch counterparts,
they require expensive parallel hardware;
as a result the performance-to-price ratio remains unchanged.
Moreover,
parallel algorithms require additional resources to communicate and synchronize the topic distributions from distributed cores/processors,
e.g.,
$1024$ processors can achieve only around $700$ times speedup with a parallel efficiency of approximately $0.7$~\cite{Newman:09}.
Finally,
it is difficult to write and debug parallel algorithms on distributed systems.

\section{Related Work} \label{s3}

The probabilistic topic modeling task can be interpreted as a labeling problem,
in which the objective is to assign a set of thematic topic labels,
$\mathbf{z}_{W \times D} = \{z^k_{w,d}\}$,
to explain the observed elements in document-word matrix,
$\mathbf{x}_{W \times D} = \{x_{w,d}\}$.
The notations $1 \le w \le W$ and $1 \le d \le D$ are the word index in vocabulary and the document index in corpus.
The notation $1 \le k \le K$ is the topic index.
The nonzero element $x_{w,d} \ne 0$ denotes the number of word counts at the index $\{w,d\}$.
For each word token $x_{w,d,i} = \{0,1\}, x_{w,d} = \sum_i x_{w,d,i}$,
there is a topic label $z^k_{w,d,i} = \{0,1\}, \sum_{k=1}^K z^k_{w,d,i} = 1, 1 \le i \le x_{w,d}$.
Integrating out the document-specific topic proportions $\theta_d(k)$ and topic distribution over vocabulary words $\phi_w(k)$ in LDA
yields the joint probability of the collapsed LDA~\cite{Heinrich:08},
\begin{align} \label{joint}
p(\mathbf{x},\mathbf{z};\alpha,\beta) \propto
&\prod_{d}\prod_{k}\Gamma\biggl(\sum_{w,i} x_{w,d,i}z^k_{w,d,i} + \alpha\biggr) \times \notag \\
&\prod_{w}\prod_{k}\Gamma\biggl(\sum_{d,i} x_{w,d,i}z^k_{w,d,i} + \beta\biggr) \times \notag \\
&\prod_{k}\Gamma\biggl(\sum_{w,d,i} x_{w,d,i}z^k_{w,d,i} + W\beta\biggr)^{-1},
\end{align}
where $\Gamma(\cdot)$ is the gamma function,
and $\{\alpha, \beta\}$ are fixed symmetric Dirichlet hyperparameters~\cite{Griffiths:04}.

\subsection{Message Passing}

The approximate BP for LDA~\cite{Zeng:11} calculates the conditional probability,
$\mu_{w,d,i}(k) = p(z^k_{w,d,i}=1|\mathbf{z}^k_{-(w,d,i)},\mathbf{x})$,
referred to as {\em message},
which can be normalized efficiently by local computation, i.e.,
$\sum_{k=1}^K \mu_{w,d,i}(k) = 1, 0 \le \mu_{w,d,i}(k) \le 1$.
The notation $\mathbf{z}^k_{-(w,d,i)}$ represents all topic labels except the topic label for the word token $x_{w,d,i}$.
According to the Bayes' rule,
we get
\begin{align} \label{marginal}
p(z^k_{w,d,i}=1|\mathbf{z}^k_{-(w,d,i)}, \mathbf{x}) =
\frac{p(\mathbf{z}^k, \mathbf{x})}{p(\mathbf{z}^k_{-(w,d,i)}, \mathbf{x})}.
\end{align}
Substituting the joint probability~\eqref{joint} into~\eqref{marginal},
we get
\begin{align} \label{marginal2}
\mu_{w,d,i}(k) \propto &\frac{\Gamma(\sum_{w,i} x_{w,d,i}z^k_{w,d,i} + \alpha)}{\Gamma(\sum_{w,-i} x_{w,d,i}z^k_{w,d,i} + \alpha)} \times \notag \\
&\frac{\Gamma(\sum_{d,i} x_{w,d,i}z^k_{w,d,i} + \beta)}{\Gamma(\sum_{d,-i} x_{w,d,i}z^k_{w,d,i} + \beta)} \times \notag \\
&\frac{\Gamma(\sum_{-(w,d,i)} x_{w,d,i}z^k_{w,d,i} + W\beta)}{\Gamma(\sum_{w,d,i} x_{w,d,i}z^k_{w,d,i} + W\beta)}.
\end{align}
Using the property $\Gamma(a+1)/\Gamma(a) = a$,
we can cancel the shared terms in both numerator and denominator of~\eqref{marginal2}
because the difference between numerator and denominator in Gamma functions is one token $x_{w,d,i}z^k_{w,d,i}$,
\begin{align} \label{GS}
&\mu_{w,d,i}(k) \propto \notag \\
&\frac{(\sum_{w,-i} x_{w,d,i}z^k_{w,d,i} + \alpha) \times (\sum_{d,-i} x_{w,d,i}z^k_{w,d,i} + \beta)}
{\sum_{-(w,d,i)} x_{w,d,i}z^k_{w,d,i} + W\beta},
\end{align}
which is exactly the same with the conditional probability of a topic label given an observed word token in the GS algorithm~\cite{Griffiths:04}.

Although different word tokens $x_{w,d,i}$ at the same index $\{w,d\}$ have different topic labels $z^k_{w,d,i}$,
they may have almost the same message.
Hence,
we approximate messages at word tokens by messages at word indices,
$\mu_{w,d,i}(k) \approx \mu_{w,d}(k), 1 \le i \le x_{w,d}$.
We define the soft topic label for the word index $\{w,d\}$ by $z_{w,d}^k = \sum_{i=1}^{x_{w,d}} z^k_{w,d,i}x_{w,d,i}/x_{w,d}$,
which is an average topic labeling configuration over all word tokens at index $\{w,d\}$.
In this way,
$\sum_k z_{w,d}^k = 1, 0 \le z_{w,d}^k \le 1$ can be viewed as the topic labeling likelihoods for the word index,
which can be approximated by the message at word index,
i.e.,
$z_{w,d}^k \approx \mu_{w,d}(k)$.
Thus,
we have the second approximation,
$x_{w,d}\mu_{w,d}(k) \approx \sum_i x_{w,d,i}z^k_{w,d,i}$.

Based on these approximations,
we get the following approximate message update equation from~\eqref{GS},
\begin{align} \label{message}
\mu_{w,d}^{t+1}(k) \propto
&\frac{ [\boldsymbol{\mu}_{-w,d}^t(k) + \alpha]\times[\boldsymbol{\mu}_{w,-d}^t(k) + \beta]}{\boldsymbol{\mu}_{-(w,d)}^t(k) + W\beta},
\end{align}
where $0 \le t \le T$ is the iteration index,
and $\boldsymbol{\mu}^t_{-w,d}(k) = \sum_{-w} x_{w,d}\mu^t_{w,d}(k)$,
$\boldsymbol{\mu}^t_{w,-d}(k) = \sum_{-d} x_{w,d}\mu^t_{w,d}(k)$,
and
$\boldsymbol{\mu}^t_{-(w,d)}(k) = \sum_{-(w,d)} x_{w,d}\mu^t_{w,d}(k)$.
The notations $-w$ and $-d$ denote all word indices except $w$ and all document indices except $d$.
Rather than subtracting one word token $x_{w,d,i}z_{w,d,i}^k$ in~\eqref{GS},
we subtract a relatively small number $x_{w,d}z_{w,d}^k$ in both numerator and denominator in~\eqref{message},
which does not affect the topic modeling performance in practice~\cite{Zeng:11}.
After updating~\eqref{message},
we normalize each message locally by the factor $Z = \sum_{k=1}^K \mu_{w,d}(k)$,
which requires additional $K$ iterations.
Messages are passed until convergence or the maximum number of iterations $T$ is reached.
Using the converged messages,
we can estimate the document-specific topic proportion $\theta_d$ and
the topic distribution over vocabulary $\phi_w$ as follows~\cite{Zeng:11},
\begin{gather}
\label{thetad}
\theta_d(k) = \frac{\boldsymbol{\mu}_{\cdot,d}(k) + \alpha}{\sum_k [\boldsymbol{\mu}_{\cdot,d}(k) + \alpha]}, \\
\phi_w(k) = \frac{\boldsymbol{\mu}_{w,\cdot}(k) + \beta}{\sum_w [\boldsymbol{\mu}_{w,\cdot}(k) + \beta]}.
\label{phiw}
\end{gather}

Batch LDA algorithms may be interpreted within the message passing or expectation-maximization (EM) framework~\cite{Zeng:11,Zeng:13}.
For example,
VB~\cite{Blei:03} is a variational message passing method~\cite{Winn:05} that
uses a set of factorized variational distributions
to approximate the objective joint distribution by minimizing the Kullback-Leibler (KL) divergence between them.
Because there is always a gap between the variational distribution and the true joint distribution,
VB causes biases when updating messages~\eqref{message}.
GS~\cite{Griffiths:04} randomly samples a topic label from the message~\eqref{GS},
which is theoretically more accurate but converges practically with more number of iterations than VB.
In contrast,
BP does not draws random samples from messages~\eqref{GS} but pass messages~\eqref{message} directly to the next iteration.
As a result,
BP retains all uncertainties encoded in the message for a higher topic modeling accuracy than GS.
In addition,
the deterministic message passing process of BP converges much faster than GS~\cite{Zeng:11}.
BP resembles CVB0~\cite{Asuncion:09,Asuncion:10} but with two main distinctions.
First,
BP infers messages based on word indices $\{w,d\}$,
while CVB0 infers messages based on word tokens $\{w,d,i\}$.
Although CVB0 and BP can achieve almost the same topic modeling accuracy,
CVB0 runs slower than BP because the number of word tokens is often significantly larger
than the number of non-zero ($NNZ$) elements in document-word matrix,
i.e.,
$\sum_{w,d}x_{w,d} \gg NNZ$.
Second,
BP for LDA is an approximation to the sum-product algorithm~\cite{Kschischang:01}
based on the factor graph representation of the collapsed LDA~\cite{Zeng:11},
while CVB0 is the zero-order mean-field approximation to the collapsed variational Bayes (CVB) algorithm~\cite{Teh:07}.
More detailed discussions and experimental comparisons between BP
and other batch LDA algorithms can be found in~\cite{Zeng:11,Asuncion:09}.

\subsection{Hypergraph Representation of Collapsed LDA}

\begin{figure*}[t]
\centering
\includegraphics[width=0.7\linewidth]{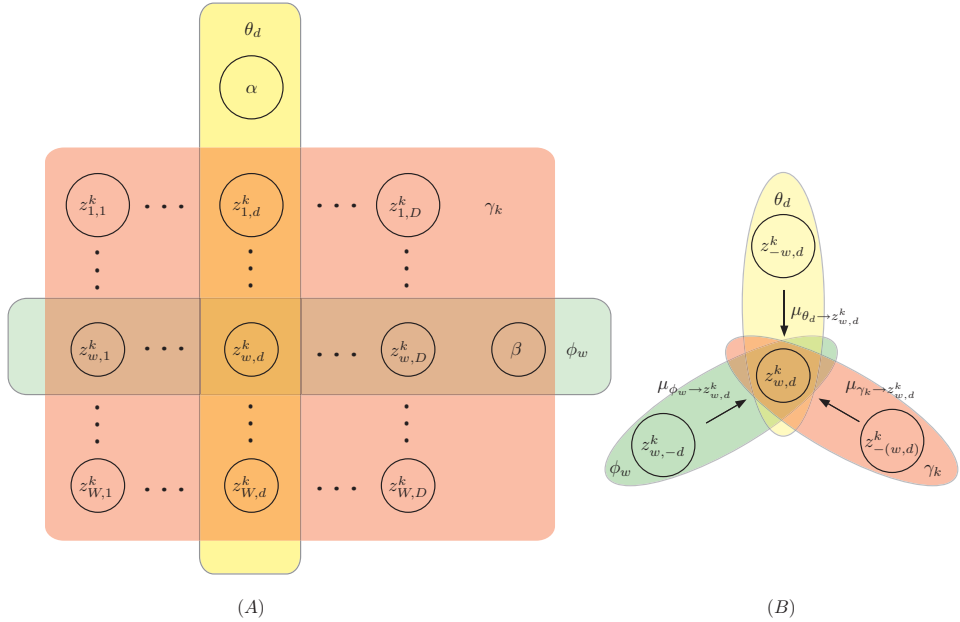}
\caption{(A) Hypergraph representation for the collapsed LDA and (B) message passing on hypergraph.}
\label{hyper}
\end{figure*}

Fig.~\ref{hyper}A shows the hypergraph for the joint probability~\eqref{joint}.
There are three types of hyperedges $\{\theta_d, \phi_w, \gamma_k\}$ denoted by the yellow, green and red rectangles,
respectively.
For each column of $\mathbf{z}_{W \times D}$,
the hyperedge $\theta_d$,
which corresponds to the first term in~\eqref{joint},
connects the variable $z_{w,d}^k$ with the subset of variables $\mathbf{z}_{-w,d}^k$ and the hyperparameter $\alpha$ within the document $d$.
For each row of $\mathbf{z}_{W \times D}$,
the hyperedge $\phi_w$,
which corresponds to the second term in~\eqref{joint},
connects the variable $z_{w,d}^k$ with the subset of variables $\mathbf{z}_{w,-d}^k$ and the hyperparameter $\beta$ at the same word $w$ in the vocabulary.
Finally,
the hyperedge $\gamma_k$,
which corresponds to the third term in~\eqref{joint},
connects the variables $z_{w,d}^k$ with all other variables $\mathbf{z}_{-(w,d)}^k$ and the hyperparameter $\beta$ on the same topic $k$.
The notations $\mathbf{z}^k_{-w,d}$, $\mathbf{z}^k_{w,-d}$
and $\mathbf{z}^k_{-(w,d)}$ represent all neighboring labeling configurations through three types of hyperedges $\{\theta_d, \phi_w, \gamma_k\}$,
respectively.
For a better illustration in Fig.~\ref{hyper},
we do not show the hyperedge that gives $\sum_k z_{w,d}^k = 1$.
Fig.~\ref{hyper}B explains the message passing equation~\eqref{message}.
We see that the variable $z_{w,d}^k$ is influenced by three subsets of neighboring variables
$\{\mathbf{z}_{-w,d}^k, \mathbf{z}_{w,-d}^k, \mathbf{z}^k_{-(w,d)}\}$,
corresponding to three types of messages
$\mu_{\theta_d \rightarrow z^k_{w,d}}$,
$\mu_{\phi_w \rightarrow z^k_{w,d}}$,
and $\mu_{\gamma_k \rightarrow z^k_{w,d}}$
through three types of hyperedges $\{\theta_d, \phi_w, \gamma_k\}$, respectively.
Indeed,
the message passing process is an approximation of the standard sum-product BP algorithm.
More detailed approximation steps can be found in~\cite{Zeng:11}.

\subsection{Residual Belief Propagation}

The synchronous BP~\cite{Zeng:11} updates all messages simultaneously at iteration $t$ based on the messages at previous iteration $t-1$.
The asynchronous schedule updates the message of each variable in a given order;
and the updated messages in turn are used to update other messages of neighboring variables at each iteration $t$.
We use RBP~\cite{Elidan:06} for the informed schedule of asynchronous message passing.
The basic idea of RBP is to select the best update order based on the residuals $r_{w,d}(k)$
defined as the $p$-norm of difference between two $K$-tuple message vectors at successive iterations.
Here,
we choose the $L_1$ norm with $p=1$,
\begin{align} \label{RBP}
r_{w,d}(k) = x_{w,d}|\mu^t_{w,d}(k) - \mu^{t-1}_{w,d}(k)|,
\end{align}
where $x_{w,d}$ is the number of word counts.
If we sequentially update messages in descending order of $r_{w,d}=\sum_k r_{w,d}(k)$ at each iteration,
the RBP algorithm theoretically converges faster or more often to a fixed point than synchronous BP~\cite{Elidan:06}.
Intuitively,
RBP always updates messages with the largest residuals first,
which in turn influence their neighboring messages efficiently.
So,
RBP accelerates the convergence speed of synchronous BP.
More details on theoretical and experimental analysis of RBP's convergence property can be found in~\cite{Elidan:06,Zeng:12a}.

\section{Active Belief Propagation} \label{s4}

In this section,
we propose ABP within the RBP framework.
First,
we describe in detail how to actively select the subset of documents and topics at each iteration for fast topic modeling.
Second,
we explain the effectiveness of ABP by Zipf's law.
Finally,
we discuss ABP's relationship to previous algorithms.

\subsection{The ABP Algorithm}

At each training iteration,
ABP actively selects the subset of documents $\lambda_dD$ from corpus for message updating and passing,
where the parameter $\lambda_d \in (0,1]$ controls the scanned corpus proportion.
For each document,
ABP searches the subset of topic space $\lambda_kK$ for message updating and passing,
where the parameter $\lambda_k \in (0,1]$ controls the proportion of the searched topic space;
that is,
ideally ABP consumes a faction ($\lambda_d\lambda_k$) of training time required by BP at each iteration.

Based on~\eqref{RBP},
we define the residuals at topics for specific documents as
\begin{align} \label{rtopic}
r_d(k) = \sum_w r_{w,d}(k),
\end{align}
and accumulate all topic residuals at documents as
\begin{align} \label{rdoc}
r_d = \sum_k r_{d}(k).
\end{align}
After each iteration,
we sort $r_d(k)$ in descending order for all topics,
and select the subset topics $\lambda_kK$ with the largest residuals $r_d(k)$ for each document.
We also sort $r_d$ in descending order for all documents,
and select the subset documents $\lambda_dD$ with the largest residuals $r_d$.
In the following iteration,
we update and pass only messages for the subset documents $\lambda_dD$ and the subset topics $\lambda_kK$,
and thus save enormous training time in each loop.
At a negligible computational cost,
the sorted residuals can be updated during message passing process.
The computational cost of initial sorting~\eqref{rtopic} and~\eqref{rdoc} using the standard ``quick sort" algorithm
are $\mathcal{O}(K\log K)$ and $\mathcal{O}(D\log D)$.
If the successive residuals are in nearly sorted order,
only a few swaps will restore the sorted order by the standard ``insertion sort" algorithm,
thereby saving lots of sorting time.\footnote{In practice,
partial sorting for $k$ largest elements is more efficient and retains almost the same performance.}

For the selected $\lambda_kK$ topics,
we need to normalize the local messages.
At the first iteration $t=1$,
ABP runs the same as BP that updates and normalizes all messages for all topics.
In the successive iterations $t \ge 2$,
ABP actively selects the subset $\lambda_k K$ topics based on residuals for message updating and passing.
We normalize the local messages for the selected topics $k \in \lambda_k K$ by
\begin{align} \label{topicmessage}
\hat{\mu}^t_{w,d}(k) = \frac{\mu^t_{w,d}(k)}{\sum_{k}\mu^t_{w,d}(k)} \times \sum_k\hat{\mu}_{w,d}^{t-1}(k), k \in \lambda_k K,
\end{align}
where $\hat{\mu}_{w,d}^{t-1}$ is the normalized message in the previous iteration,
$\hat{\mu}^t_{w,d}(k)$ is the normalized message in the current iteration,
and $\mu^t_{w,d}(k)$ is the unnormalized message updated according to~\eqref{message}.
In this way,
we need only $\lambda_k K$ iterations to avoid calculating
the normalization factor $Z$ with $K$ iterations.

\begin{figure}[t]
\centering
\includegraphics[width=0.9\linewidth]{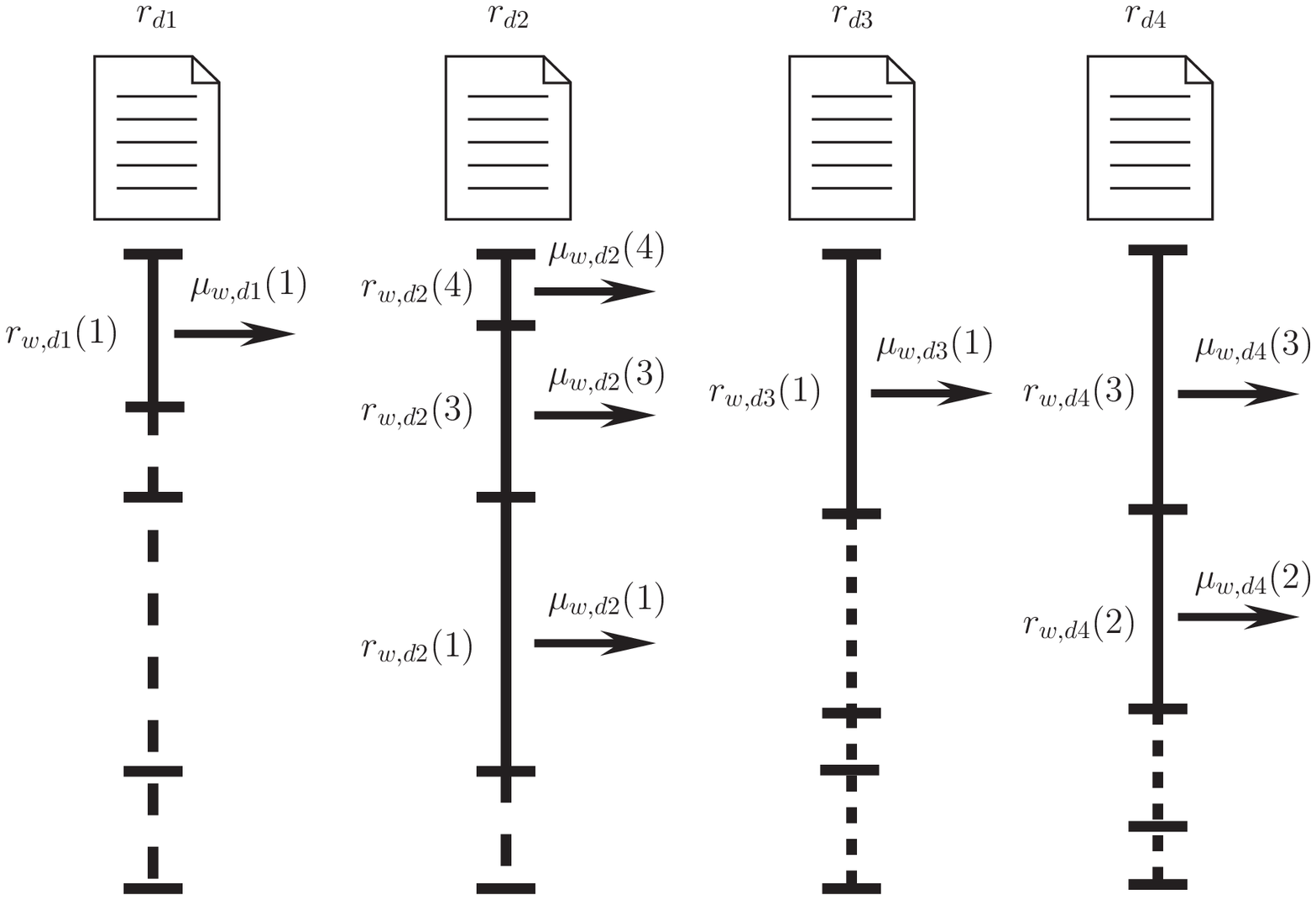}
\caption{An example of active message passing.}
\label{residual}
\end{figure}

Fig.~\ref{residual} shows an example of the active message passing process.
The solid and dashed segments denote normalized messages in proportion.
The solid segments and arrows denote those selected messages for updating and passing,
while the dashed segments denote those unchanged messages.
Suppose that documents $d1$, $d2$, $d3$ and $d4$ have largest residuals $r_{d1}$, $r_{d2}$, $r_{d3}$ and $r_{d4}$.
We select these four documents in the subset $\lambda_dD$ for active message passing.
For the document $d1$,
only $r_{w,d1}(1)$ is in the subset $\lambda_kK$ so that the message $\mu_{w,d1}(1)$ is updated and passed.
For the document $d2$,
three residuals $r_{w,d2}(4)$, $r_{w,d2}(3)$ and $r_{w,d2}(1)$ are ranked in the subset $\lambda_kK$,
and thus the messages $\mu_{w,d2}(4)$, $\mu_{w,d2}(3)$ and $\mu_{w,d2}(1)$ are updated and passed.
Similar rules are applied to documents $d3$ and $d4$.
Because only the subset of messages are updated and passed,
ABP consumes significantly less computation than BP in each loop.

We dynamically refine and sort residuals at each iteration.
Due to the contraction effect~\cite{Elidan:06},
the largest residuals in the previous iterations may be ranked lower in the subsequent iterations.
As a result,
those unchanged messages will be updated and passed as indicated by the dashed segments in Fig.~\ref{residual}.
In this sense,
ABP retains almost the same message information as BP.

Within the RBP framework~\cite{Elidan:06},
ABP will converge to a unique fixed point by satisfying the following two assumptions:
\begin{enumerate}
\item
The message update equation~\eqref{message} is a contraction mapping $f$.
\item
For every message $\mu_{w,d}(k)$,
there is a finite time interval in which the message update~\eqref{message} is executed at least once.
\end{enumerate}
The first assumption is often satisfied in most existing message passing algorithms based on factor graphs or clustered graphs~\cite{Elidan:06}.
With Fig.~\ref{residual},
we explain that the second assumption is often satisfied in practice due to the contraction effects.

\begin{figure}[t]
\centering
\includegraphics[width=1\linewidth]{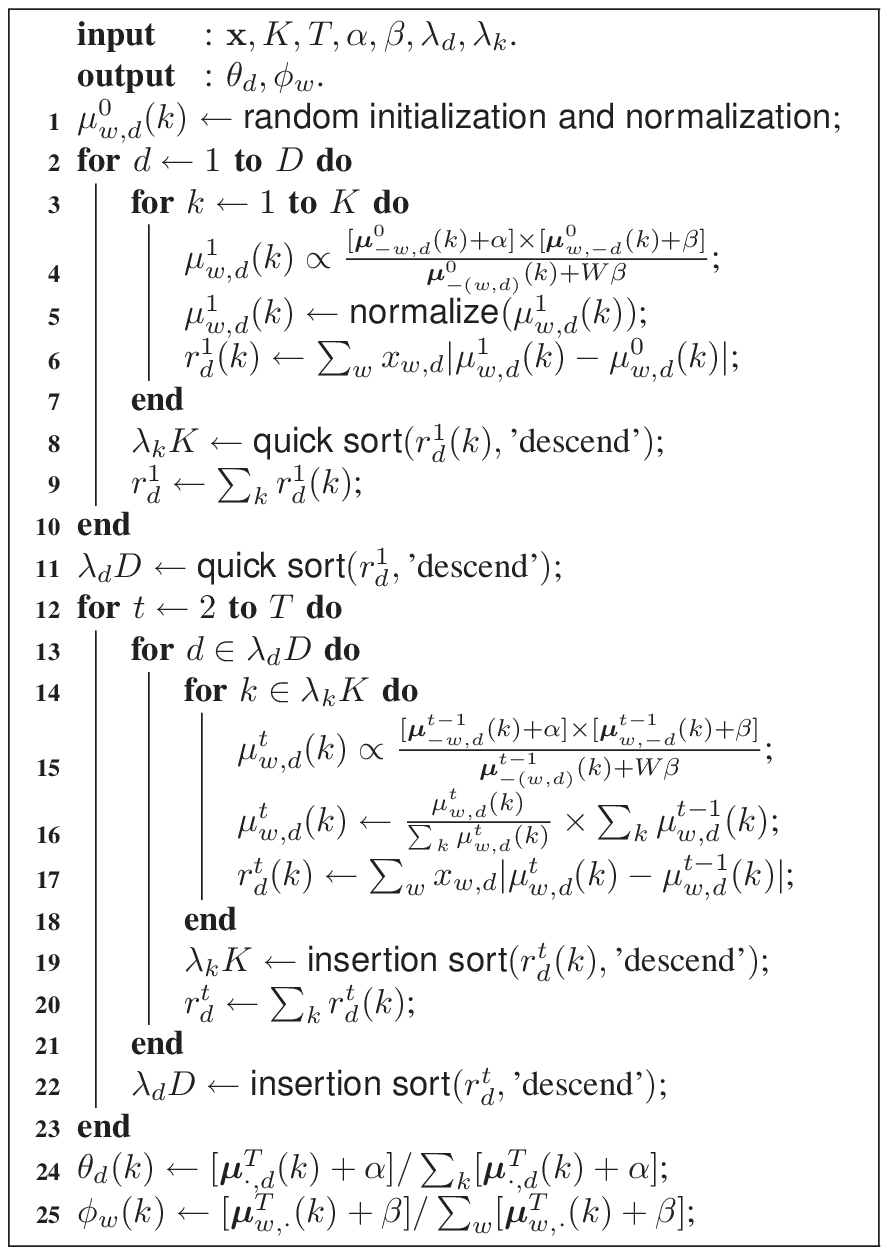}
\caption{The ABP algorithm.}
\label{code}
\end{figure}

Fig.~\ref{code} summarizes the ABP algorithm.
When $t=1$,
ABP scans the entire corpus and searches the complete topic space,
and in the meanwhile computes and sorts residuals by the ``quick sort" algorithm.
The purpose of this initial sweep is to calculate and store all residuals~\eqref{rtopic} and~\eqref{rdoc}.
For $2 \le t \le T$,
ABP actively selects the subset documents $\lambda_dD$ and the subset topics $\lambda_kK$ for message updating and passing.
At the end of each iteration,
ABP dynamically refines and sorts residuals by the ``insertion sort" algorithm.
It terminates when the maximum number of iterations $T$ is reached or the convergence condition is satisfied.

The time complexity of ABP is $\mathcal{O}(\lambda_d\lambda_kKDT)$,
where $K$ is the number of topics, $D$ the number of documents, and $T$ the number of iterations.
The space complexity of ABP is $\mathcal{O}(KD)$,
which is the same as the conventional BP or other batch LDA algorithms.
Due to memory limitation for all batch LDA algorithms,
it is hard to fit massive corpora with a large number of topics into the memory of a common desktop computer (for example, $3$G RAM).
We have two straightforward strategies to address the memory problem.
First,
we may integrate ABP with online stochastic optimization~\cite{Bottou:04} or extend ABP on the blackboard parallel architecture~\cite{Smola:10}.
Similar to other online and parallel LDA algorithms,
online ABP requires a fixed memory requirement for each mini-batch data,
while parallel ABP can use the large memory space from clusters of workstations.
Second,
we may combine ABP with the block optimization framework~\cite{Yu:10,Zeng:12b} that
sequentially reads blocks of data from the hard disk into memory for optimization.
Nevertheless,
these extensions of ABP are nontrivial at present,
and to be studied in our future work such as~\cite{Yan:12,Zeng:13}.

Here we show an alternative implementation of ABP.
For massive corpora,
initially sorting $r_d$ in~\eqref{rdoc} requires at most a computational complexity of $\mathcal{O}(D\log D)$.
This cost is very high when $D$ is very large,
for example,
$D=10^8$.
Alternatively,
we may define residuals at the fixed vocabulary,
\begin{gather} \label{rtopic2}
r_w(k) = \sum_d r_{w,d}(k),
\end{gather}
and
\begin{gather} \label{rvoc}
r_w = \sum_k r_{w}(k).
\end{gather}
We may sort $r_w$ with a significantly less computational cost of $\mathcal{O}(W\log W)$
because $W$ is often a fixed value whereas $D$ is a huge number for a massive corpora.
In this case,
ABP actively selects the subset of vocabulary words,
$\lambda_w W, \lambda_w \in (0,1]$,
for topic modeling at each iteration.
Due to page limitation of this paper,
we do not show the experimental results of this alternative implementation.
Interested readers can find related source codes of this implementation in~\cite{Zeng:12}.

\subsection{Document Residuals Follow Zipf's Law}

\begin{figure}[t]
\centering
\includegraphics[width=0.8\linewidth]{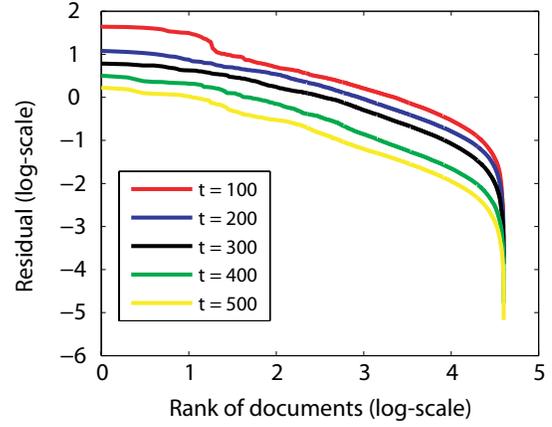}
\caption{The residual follows Zipf's law at different iterations $t \in \{100, 200, 300, 400, 500\}$ on the ENRON data set.}
\label{zipf}
\end{figure}

The residual~\eqref{RBP} reflects the magnitude of the gradient descent to maximize~\eqref{joint}.
The larger the residual the larger the gradient descent.
Since ABP updates the selected subset of messages with the largest residuals,
it in effect optimizes~\eqref{joint} using the largest gradient descents.
According to the stochastic optimization theory~\cite{Bottou:book},
the subset of the gradient descents can perform well as the entire batch gradient descents when the data set is very large.
Using the ENRON data set in Table~\ref{dataset} with $K=10$,
we show the log-log plot of the document residuals~\eqref{rdoc} relative to the rank of documents based on~\eqref{rdoc} in Fig.~\ref{zipf}.
Overall,
the residuals are reduced with the increase of the training iterations $t \in \{100,200,300,400,500\}$,
which indicates that the batch gradient descents are becoming smaller and smaller owing to convergence.
Approximately,
the residuals follow Zipf's law~\cite{Zipf:book} during training,
\begin{align}
\log(r_d) \approx a\log(rank) + b,
\end{align}
where $a$ and $b$ are constants.
Zipf's law implies that small residuals are extremely common,
whereas large residuals are extremely rare.
More specifically,
we find that top $20\%$ documents are responsible for almost $80\%$ residuals during training,
which seems to follow the Pareto principle or the $80$-$20$ rule\footnote{\url{http://en.wikipedia.org/wiki/Pareto_principle}},
stating that roughly $80\%$ of the effects come from $20\%$ of the causes.
Therefore,
ABP can efficiently use only $20\%$ documents to obtain $80\%$ gradient descents to optimize~\eqref{joint} at each training iteration,
leading to a fast speed while achieving a comparable topic modeling accuracy.
Since many other natural data sets approximately follow Zipf's law,
the basic idea of ABP can be extended to speeding up other learning algorithms for these data sets.

\subsection{Relationship to Previous Algorithms}

Unlike the parallel LDA algorithms~\cite{Newman:09,Mimno:12},
ABP is a batch LDA algorithm that uses only the single processor/core.
It selects only a mini-batch of documents from the entire corpus based on residuals for message updating and passing at each iteration.
From this perspective,
ABP is similar to OVB~\cite{Hoffman:10} but with three main distinctions.
First,
OVB sequentially read each mini-batch from the data stream without selection.
Second,
OVB discards each mini-batch after one look.
Finally,
ABP can converge to a local optimum of the BP's objective function,
while OVB requires specific parameter settings to achieve this goal.
SOI~\cite{Mimno:12},
an extension of OVB,
randomly re-use previous seen mini-batches,
so it does not schedule those mini-batches for a better performance.

ABP differs from RVB~\cite{Wahabzada:11} though they both use the residual-based dynamical scheduling techniques.
First,
ABP is a batch algorithm based on the BP inference while RVB is derived from the OVB~\cite{Hoffman:10} inference.
Second,
ABP uses a more efficient sorting method while RVB uses a relatively complicated sampling technique for dynamical scheduling.
Third,
ABP can dynamically schedule documents,
vocabulary words,
and topics for the maximum speedup effects,
while RVB can schedule only mini-batches of documents.
Finally,
RVB uses only the lower bound of residuals~\eqref{rdoc},
which may lower the scheduling efficiency.

The FGS and SGS algorithms~\cite{Porteous:08,Yao:09} are two important improvements over the batch GS algorithm when $K$ is very large.
The basic idea of FGS is to combine both message update and sampling process together by introducing an upper bound $\hat{Z}$ for the normalization factor $Z$.
Similar to FGS,
we may define an upper bound and pass only the largest segments $\mu_{w,d}(k)/\hat{Z}$ in BP to reduce the computation cost.
However,
this strategy does not work because we will not be able to update and pass those short segments (small messages) illustrated in Fig.~\ref{residual},
leading to serious loss of information.
Although such a strategy is faster,
its accuracy is much lower than BP.
So,
we formulate ABP within the RBP framework~\cite{Elidan:06},
which dynamically refines and sorts residuals to determine the best subsect $\lambda_kK$ topic space at each iteration.

\section{Experiments} \label{s5}

\begin{table}
\centering
\caption{Statistics of four document data sets.}
\begin{tabular}{|l|l|l|l|l|} \hline
Data sets &$D$       &$W$     &$N_d$      &$W_d$      \\ \hline \hline
NIPS      &$1500$    &$12419$ &$311.8$    &$217.6$    \\ \hline
ENRON     &$39861$   &$28102$ &$160.9$    &$93.1$     \\ \hline
NYTIMES   &$15000$   &$84258$ &$328.7$    &$230.2$    \\ \hline
PUBMED    &$80000$   &$76878$ &$68.4$     &$46.7$    \\
\hline\end{tabular}
\label{dataset}
\end{table}

Our experiments aim to verify the accelerating effects of ABP compared with that of
other state-of-the-art batch LDA algorithms including VB~\cite{Blei:03}, GS~\cite{Griffiths:04}, FGS~\cite{Porteous:08},
SGS~\cite{Yan:09},
CVB0~\cite{Asuncion:09} and synchronous BP~\cite{Zeng:11}.
We use four publicly available document data sets~\cite{Porteous:08}:
NIPS, ENRON, NYTIMES and PUBMED.
Previous studies~\cite{Porteous:08} have revealed that the speedup effect is relatively insensitive to the number of documents in the corpus.
Because of the memory limitation for batch LDA algorithms,
we randomly select $15000$ documents from the original NYTIMES data set,
and $80000$ documents from the original PUBMED data set for experiments.
Table~\ref{dataset} summarizes the statistics of four data sets,
where $D$ is the total number of documents in the corpus,
$W$ is the number of words in the vocabulary,
$N_d$ the average number of word tokens per document,
and $W_d$ the average number of word indices per document.

We randomly partition each data set into halves with one for training set and the other for test set.
We calculate the training perplexity~\cite{Asuncion:09} on the training set after $500$ iterations as follows,
\begin{align} \label{training}
\mathcal{P} = \exp\Bigg\{-\frac{\sum_{w,d}
x_{w,d}\log\big[\sum_{k}\theta_d(k)\phi_w(k)\big]}
{\sum_{w,d} x_{w,d}}\Bigg\}.
\end{align}
Usually,
the training perplexity will decrease with the increase of the number of training iterations.
The algorithm often converges if the change of training perplexity at successive iterations is less than a predefined threshold.
In our experiments,
we set the threshold to one because the decrease of training perplexity will be very small after satisfying this threshold.

The predictive perplexity for the unseen test set is computed as follows~\cite{Asuncion:09}.
On the training set,
we estimate $\phi$ from the same random initialization after $500$ iterations.
For the test set,
we randomly partition each document into $80\%$ and $20\%$ subsets.
Fixing $\phi$,
we estimate $\theta$ on the $80\%$ subset by the training algorithms from the same random initialization after $500$ iterations,
and then calculate the predictive perplexity on the rest $20\%$ subset,
\begin{align} \label{test}
\mathcal{P}=\exp\Bigg\{-\frac{\sum_{w,d}
x_{w,d}^{20\%}\log\big[\sum_{k}\theta_d(k)\phi_w(k)\big]}
{\sum_{w,d} x_{w,d}^{20\%}}\Bigg\},
\end{align}
where $x_{w,d}^{20\%}$ denotes word counts in the the $20\%$ subset.
The lower predictive perplexity represents a better generalization ability.

For all data sets,
we fix the same hyperparameters $\alpha = 2/K$ and $\beta = 0.01$~\cite{Porteous:08}.
The CPU time per iteration is measured after sweeping the entire data set.
We report the average CPU time per iteration after $T = 500$ iterations,
which practically ensures that GS and FGS converge in terms of training perplexity.
For a fair comparison,
we use the same random initialization to examine all algorithms with $500$ iterations.
We implement all algorithms based on the MEX C++/MATLAB/Octave platform,
where GS/FGS C++ source codes are the same as those in~\cite{Porteous:08},
and SGS source codes are the same as those in Mallet package.\footnote{\url{http://mallet.cs.umass.edu/}}
To repeat our experiments,
we have made all source codes and data sets publicly available~\cite{Zeng:12}.
Without the parallel implementations,
all algorithms are run on the Sun Fire X4270 M2 server with two 6-core $3.46$ GHz CPUs and $128$ GB RAMs.

\subsection{Parameters $\lambda_d$ and $\lambda_k$}

\begin{figure}[t]
\centering
\includegraphics[width=0.6\linewidth]{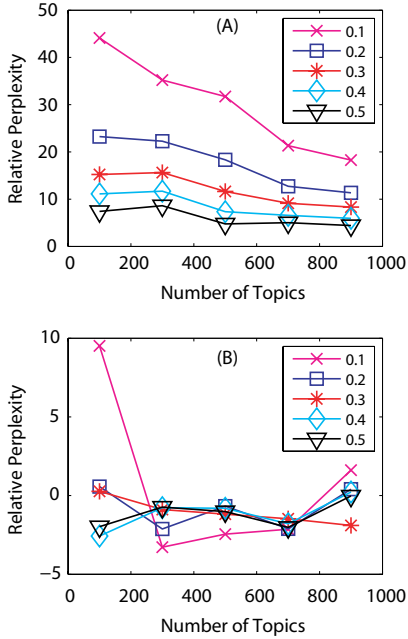}
\caption{The relative training perplexity as a function of $K$ when the parameter (A) $\lambda_d \in \{0.1,0.2,0.3,0.4,0.5\}$ and (B)
$\lambda_k \in \{0.1,0.2,0.3,0.4,0.5\}$.}
\label{parameter}
\end{figure}

First,
we examine two parameters $\lambda_d$ and $\lambda_k$ in ABP on the relatively smaller NIPS data set,
because we were wondering how these two parameters would influence the topic modeling accuracy of ABP.
The parameter $\lambda_d \in (0,1]$ controls the proportion of documents to be scanned,
and the parameter $\lambda_k \in (0,1]$ controls the proportion of topics to be searched at each iteration.
The smaller values correspond to the faster speed of ABP.
We choose the training perplexity $\mathcal{P}_{\lambda_d=1,\lambda_k=1}$ of ABP with $500$ iterations when $\lambda_d=\lambda_k=1$ as the benchmark.
The relative training perplexity is the difference between this benchmark and the ABP's training perplexity~\eqref{training}
with $500$ iterations at other parameter values,
\begin{align}
\Delta \mathcal{P} = \mathcal{P}_{\lambda_d,\lambda_k} - \mathcal{P}_{\lambda_d=1,\lambda_k=1}.
\end{align}

Fixing $\lambda_k=1$,
we change $\lambda_d$ from $0.1$ to $0.5$ with a step size $0.1$.
Fig.~\ref{parameter}A shows the relative training perplexity as a function of $K$ when $\lambda_d \in \{0.1,0.2,0.3,0.4,0.5\}$.
We see that the relative perplexity decreases when $\lambda_d$ increases,
which confirms the fact that scanning more documents at each iteration will yield the lower training perplexity under different topics $K$.
Notice that the relative perplexity decreases as $K$ increases.
This phenomenon shows that when $K$ is very large,
even scanning a small portion of documents is enough to provide a comparable topic modeling accuracy.
When $\lambda_d \ge 0.2$
we also see that the relative perplexity is less than $20$,
which in practice is a negligible difference on the common document data set.
Therefore,
$\lambda_d=0.2$ is a safe bound to guarantee a good topic modeling accuracy with a relatively faster speed.
Although a smaller $\lambda_d < 0.2$ will produce a even faster speed,
it will cause an obvious degradation of the topic modeling accuracy in terms of perplexity with a fixed number of training iterations.

Fixing $\lambda_d=1$,
we also change $\lambda_k$ from $0.1$ to $0.5$.
Fig.~\ref{parameter}B shows the relative training perplexity as a function of $K$ when $\lambda_k \in \{0.1,0.2,0.3,0.4,0.5\}$.
Surprisingly,
there is no big difference when $\lambda_k = 0.1$ and $\lambda_k = 0.5$ especially when $K \ge 300$.
This phenomenon implies that only a small proportion of topics plays a major role when $K$ is very large.
When $\lambda_k \le 0.5$,
ABP achieves even a lower perplexity value than ABP with $\lambda_k=1$.
The reason is that most documents have very sparse messages when $K$ is very large,
and thus searching the subset of topic space is enough to yield a comparable topic modeling accuracy.
Such a property as sparseness of messages has been also used to speed up topic modeling by FGS~\cite{Porteous:08} and SGS~\cite{Yao:09}.
We wonder whether $\lambda_k$ can be even smaller when $K$ is very large, e.g.,
$K \in \{1500,2000\}$.
On NIPS data set,
ABP with $\lambda_k = 0.05$ achieves $555.89$ and $542.70$ training perplexity, respectively.
In contrast,
ABP with $\lambda_k=1$ achieves $543.90$ and $533.97$ training perplexity, respectively.
The relative training perplexity is less than $2\%$.
Therefore,
it reasonable to expect that when $K$ is very large like $K \ge 2000$,
$\lambda_k K$ may be a constant, e.g., $\lambda_k K = 100$.
In this case,
the training time of ABP will be independent of $K$.
This bound $\lambda_k K = 100$ is reasonable because
usually a common word is unlikely to be associated with more than $100$ topics in practice.

Users may set different parameters $\lambda_d$ and $\lambda_k$ for different speedup effects.
To pursue the maximum speedup,
we choose $\lambda_d = \lambda_k = 0.1$ referred to as ABP1 in the rest of our experiments.
To ensure the topic modeling accuracy,
we also choose the safe bound $\lambda_d = \lambda_k = 0.2$ referred to as ABP2.
Obviously,
ABP1 is faster than ABP2 but with the lower accuracy
in terms of the training perplexity using a fixed number of training iterations.

\subsection{Predictive Perplexity}

\begin{figure*}[t]
\centering
\includegraphics[width=0.8\linewidth]{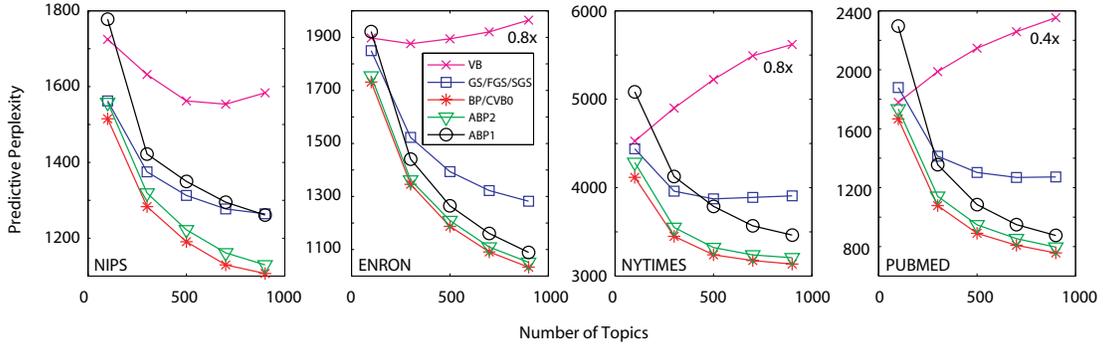}
\caption{Predictive perplexity as a function of topics $K=\{100,300,500,700,900\}$ on NIPS, ENRON, NYTIMES and PUBMED data sets.
The notation $0.8x$ denotes the predictive perplexity is multiplied by $0.8$.}
\label{prediction}
\end{figure*}

The predictive perplexity is a widely-used performance measure for the generalization ability~\cite{Blei:03,Asuncion:09,Hoffman:10},
especially for different training algorithms of LDA.

Fig.~\ref{prediction} compares the predictive perplexity as a function of $K \in \{100,300,500,700,900\}$.
GS, FGS and SGS have exactly the same topic modeling accuracy,
so that their predictive perplexity curves overlap denoted by blue squares.
On ENRON, NYTIMES and PUBMED data sets,
VB shows an obvious overfitting phenomenon,
where the predictive perplexity increases with the number of topics $K$.
Also,
VB often yields the highest predictive perplexity.
For a better visualization,
we multiply VB's predictive perplexity value by $0.8$ and $0.4$ on ENRON, NYTIMES and PUBMED data sets, respectively.
Confronted with a large number of topics on large data sets,
VB often generalizes badly to predict unseen test set.
The major reason is that VB aims to optimize an approximate variational distribution with a gap from the joint distribution~\eqref{joint}.
Our experiments show that this gap or bias is obvious in case of the massive data set containing a large number of topics.
In practice,
VB may choose the proper hyperparameters to correct this bias~\cite{Asuncion:09}.
On all data sets,
BP consistently yields the lowest predictive perplexity.
This result is consistent with~\cite{Zeng:11}.
Unlike VB,
BP decreases the predictive perplexity with the increase of the number of topics $K$ without overfitting.
Because BP directly infers the conditional posterior probability from~\eqref{joint},
it often generalizes to unseen documents well in practice.
Similarly,
CVB0 infers the messages on the word tokens instead of the word indices,
so it has almost the same predictive perplexity as BP denoted by the red star curve as shown in Fig.~\ref{prediction}.

With $\lambda_d=\lambda_k=0.2$,
ABP2 performs almost the same as BP/CVB0,
especially on ENRON and PUBMED data sets.
This result shows that scanning $20\%$ documents and searching $20\%$ topics at each iteration
is enough to provide a comparable topic modeling accuracy as BP/CVB0.
With $\lambda_d=\lambda_k=0.1$,
ABP1 provides a relatively higher predictive perplexity than BP/CVB0.
On the relatively smaller data set NIPS,
ABP1 performs even worse than VB when $K=100$.
However,
on the relatively larger data set PUBMED and when $K$ is large, e.g., $K=900$,
ABP1 shows a comparable predictive perplexity as BP/CVB0.
When compared with GS/FGS/SGS,
ABP1 is worse under different topics on the relatively smaller NIPS set,
but is much better on the relatively larger data sets: ENRON, NYTIMES and PUBMED,
when $K$ is large.
In conclusion,
ABP2 has a comparable performance as BP/CVB0,
and outperforms GS/FGS/SGS and VB on all data sets.
While ABP1 has a higher predictive perplexity than ABP2 and BP/CVB0 on all data sets,
it often outperforms GS/FGS/SGS and VB for larger data sets when the number of topics is very large ($K \ge 500$).

With the same training iterations,
ABP1 has a relatively higher predictive perplexity than BP,
which implies that ABP1 loses some topic modeling accuracy to achieve a significant speedup.
This phenomenon resembles the lossy/lossless compression for images.
To obtain a higher compression ratio,
lossy techniques will generally discard some detailed image information.
In this sense,
ABP can be viewed as a lossy learning algorithm for topic modeling with the fast speed.
By contrast,
FGS and SGS are lossless GS algorithms for topic modeling.
When $\lambda_d=\lambda_k=1$,
ABP becomes the lossless RBP algorithm for accurate topic modeling with the fast convergence speed~\cite{Zeng:12a}.
In different real-world applications,
we may have different requirements for speed and accuracy.
Users may choose different parameters $\{\lambda_d,\lambda_k\}$ to trade off between the topic modeling speed and accuracy.

\subsection{Speedup Effects}

\begin{figure*}[t]
\centering
\includegraphics[width=0.8\linewidth]{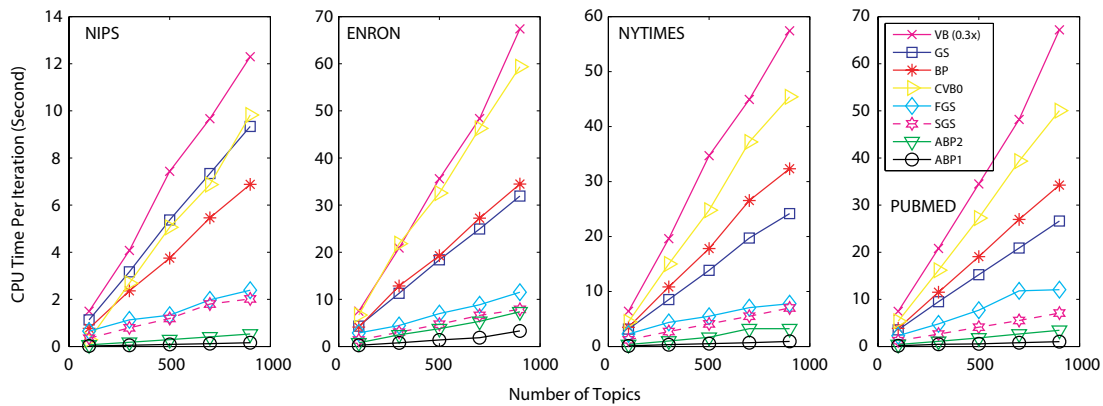}
\caption{CPU time per iteration (seconds) as a function of topics $K \in \{100,300,500,700,900\}$ on NIPS, ENRON, NYTIMES and PUBMED data sets.
The notation $0.3x$ denotes the training time is multiplied by $0.3$.}
\label{time}
\end{figure*}

\begin{figure*}[t]
\centering
\includegraphics[width=0.8\linewidth]{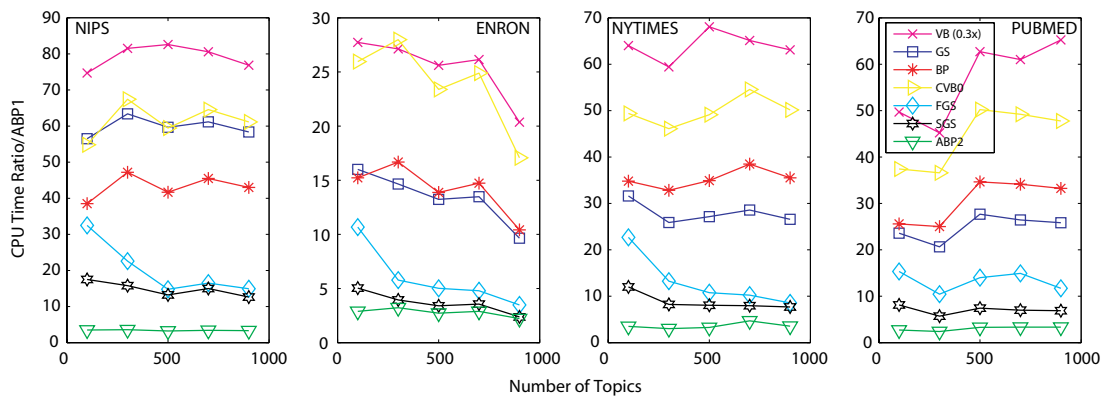}
\caption{Ratio of CPU time per iteration over ABP1 as a function of topics $K \in \{100,300,500,700,900\}$ on NIPS, ENRON, NYTIMES and PUBMED data sets.
The notation $0.3x$ denotes the CPU time is multiplied by $0.3$.}
\label{ratio1}
\end{figure*}

\begin{figure*}[t]
\centering
\includegraphics[width=0.8\linewidth]{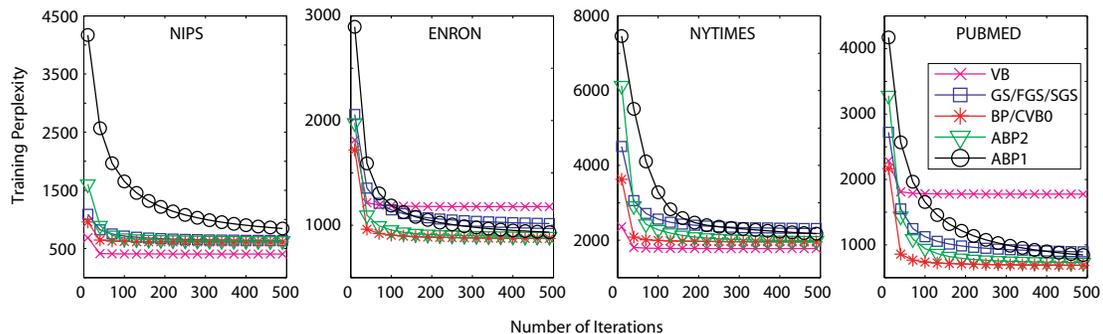}
\caption{Training perplexity as a function of the number of iterations when $K=500$ on NIPS, ENRON, NYTIMES and PUBMED data sets.}
\label{convergence}
\end{figure*}

\begin{figure*}[t]
\centering
\includegraphics[width=0.8\linewidth]{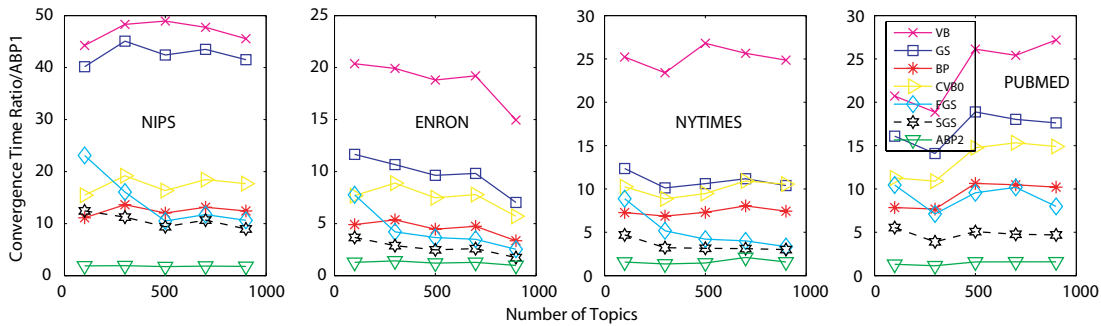}
\caption{Convergence time ratio over ABP1 as a function of topics $K \in \{100,300,500,700,900\}$ on NIPS, ENRON, NYTIMES and PUBMED data sets.}
\label{ratio2}
\end{figure*}

\begin{figure*}
\centering
\includegraphics[width=0.7\linewidth]{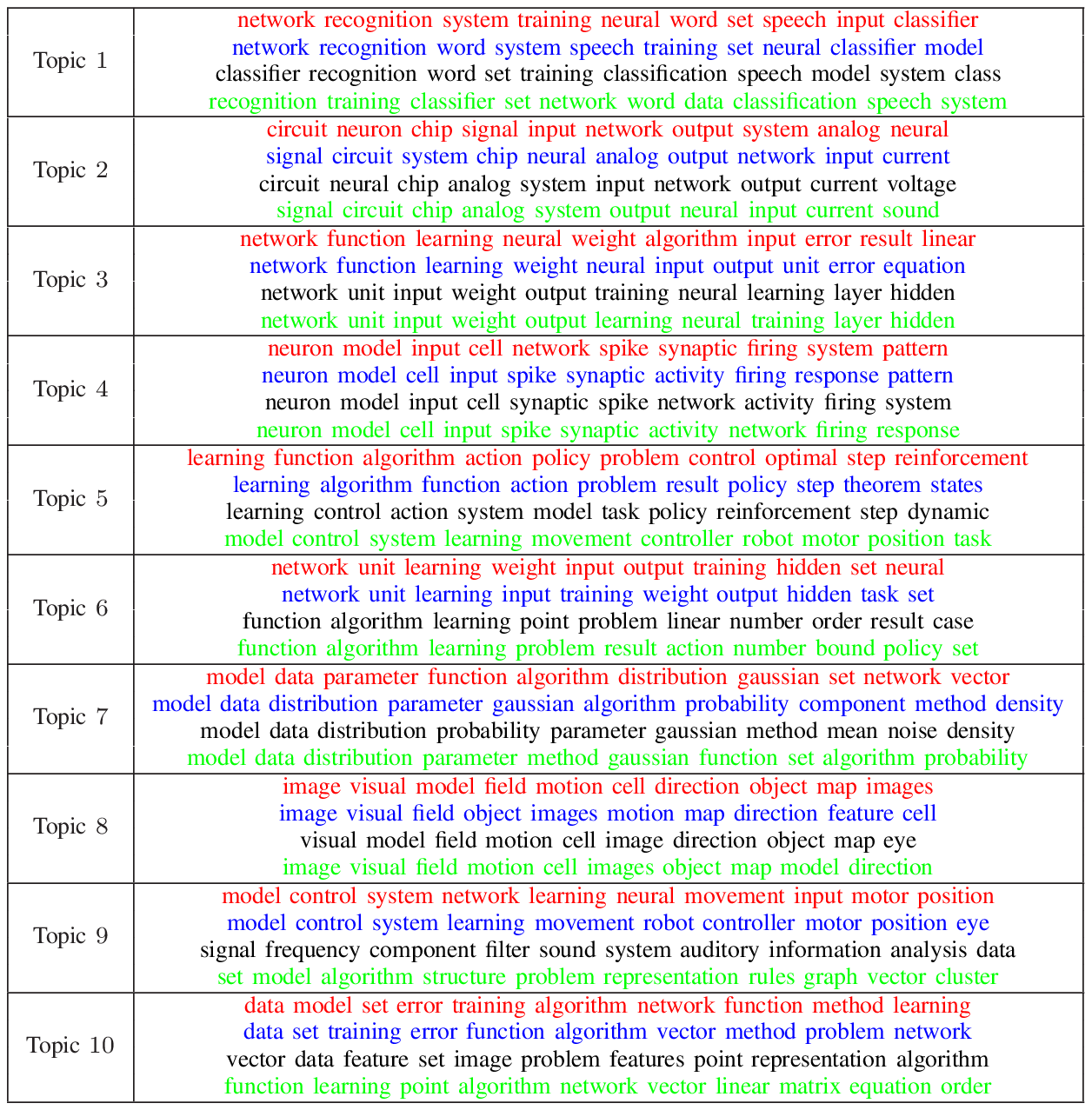}
\caption{Top ten words of ten topics on NIPS: VB (red), GS/FGS/SGS (blue), BP/CVB0 (black) and ABP (green).}
\label{topic}
\end{figure*}

Fig.~\ref{time} shows the CPU time (second) per iteration as a function of $K$ for VB, GS, FGS, SGS, BP, CVB0, ABP1 and ABP2.
The training time of all algorithms increases linearly with $K$ except FGS on the PUBMED data set.
When $K$ is very large,
FGS~\cite{Porteous:08} will locate the targeted topics more efficiently so that its training time does not increase linearly with $K$.
Because FGS does not visit all possible topic space to sample the targeted topic,
it is much faster than GS when $K$ is very large.
On four data sets,
SGS is much faster than FGS because it avoids almost all surplus computations during the sampling process.
Similar to GS,
both FGS and SGS still scan the entire corpus to sample topics for each word token.
VB consumes the longest training time due to its complicated digamma function calculation~\cite{Zeng:11}.
For a better illustration,
we multiply VB's training time by $0.3$.
This result confirms that VB is time-consuming and unsuitable for big topic modeling.

The training time of BP is shorter than that of GS when $K$ is small,
but is longer than that of GS when $K$ is large on ENRON, NYTIMES and PUBMED data sets.
The major reason is that BP requires $K$ iterations for the local message normalization,
while GS needs less than $K$ iterations because it stops visiting the rest topic space after sampling the target topic.
Because CVB0 sweeps the entire word tokens instead of word indices at each iteration,
it consumes approximately $N_d/W_d \approx 1.5$ training time of BP in Table~\ref{dataset}.
When $\lambda_d=\lambda_k = 0.1$,
ABP1 needs only $1/100$ training time of BP.
However,
as ABP1 has to sort and update residuals,
on average it consumes around $1/30$ training time of BP.
When $\lambda_d=\lambda_k = 0.2$,
ABP2 in anticipation requires four times longer training time than ABP1.
But in practice ABP1 runs around three times faster than ABP2.
Fig.~\ref{time} demonstrates that at present ABP is the fastest batch LDA algorithm.

Fig.~\ref{ratio1} shows the ratio of CPU time per iteration over ABP1 as the benchmark.
For a better illustration,
we multiply the ratio of VB by $0.3$.
We see that ABP1 runs at least $100$ times faster than VB on NIPS, NYTIMES and PUBMED data sets.
On ENRON set,
we also see that ABP1 is close to $100$ times faster than VB.
The speedup effects are significant.
For example,
if VB uses three months for training LDA,
ABP1 consumes only one day for the same task with much better generalization ability as shown in Fig.~\ref{prediction}.
ABP1 also runs around $20$ to $40$ times faster than both GS and BP.
SGS is around $1.1$ to $2$ times faster than FGS,
consistent with the result in~\cite{Yao:09}.

When compared with the relatively faster FGS and SGS,
ABP1 still runs around $10$ to $20$ times faster with the comparable predictive perplexity.
ABP2 consistently consumes $2$ to $3$ times more training time than ABP1 because it sweeps more documents and topics.
Both Figs.~\ref{prediction} and~\ref{ratio1} reconfirm that ABP
can be around $10$ to $100$ times faster than current state-of-the-art batch LDA algorithms with a comparable accuracy.

Technically,
ABP can converge to a fixed point as discussed in Section~\ref{s4}.
Fig.~\ref{convergence} shows the training perplexity as a function of number of training iterations when $K = 500$.
We see that GS/FGS/SGS converges at the same number of iterations,
and BP/CVB0 also converges at the same number of iterations with almost overlapped curves.
Usually,
GS/FGS/SGS uses $400\sim500$ iterations,
while BP/CVB0 needs $100\sim200$ iterations to achieve convergence.
Among all the algorithms,
VB converges with the least number of iterations,
which often uses around $80\sim150$ iterations for convergence.
Although VB sometimes achieves the lowest training perplexity on NIPS and NYTIMES data sets,
it provides the highest predictive perplexity on unseen test set due to overfitting in Fig.~\ref{prediction}.
ABP2 converges with the more number of iterations than BP,
which is caused by scanning the subset of corpus and searching the subset of topic space at each iteration.
In our experiments,
ABP2 uses around $300\sim500$ iterations to achieve convergence.
On NYTIMES and PUBMED data sets,
ABP2 converges with almost the same number of iterations as GS/FGS/SGS.
Furthermore,
ABP1 converges with the largest number of training iterations to reach convergence on all data sets.
In practice,
ABP1 often uses $500\sim800$ iterations for convergence,
because at each iteration it has to scan $10\%$ corpus and search $10\%$ topics for the fast speed.
However,
even if we compare the training time until convergence,
ABP1 is still the fastest algorithm.
To confirm this,
Fig.~\ref{ratio2} summarizes the training time ratio until convergence over ABP1.
We see that ABP1 still runs around $20\sim45$ times faster than VB,
$10\sim40$ times faster than GS,
and $5\sim20$ times faster than CVB0, FGS/SGS and BP when measured by the training time until convergence.

Fig.~\ref{topic} shows the top ten words of $K=10$ topics extracted by VB (red), GS/FGS/SGS (blue),
BP/CVB0 (black) and ABP (green) algorithms on the NIPS training set.
For ABP,
we choose $\lambda_d = \lambda_k = 0.5$ because $K=10$ is small.
We see that all algorithms can infer thematically meaningful topics,
where most top ten words are consistent except the slightly different word ranking.
There are two subjective measures for the interpretability of discovered topics:
word intrusion and topic intrusion~\cite{Chang:09b}.
The former is the number of word intruders in each topic,
while the latter is the number of topic intruders in each document.
According to our experience,
the word and topic intrusions are comparable among all the algorithms.
So,
given the similar interpretable topics,
we advocate ABP for fast topic modeling.

Although at each iteration ABP updates and normalizes only the subset of $K$-tuple messages~\eqref{topicmessage},
it devotes almost the equal number of iterations to different topics $K$ with two main reasons.
First,
at each iteration different documents contribute different topics according to residuals~\eqref{rtopic}.
So,
each topic will be visited by almost the equal number of times due to the LDA's clustering property~\cite{Zeng:11} across all documents.
Second,
ABP uses the dynamical scheduling of residuals shown in Fig.~\ref{code} at each iteration,
which ensures that all $K$-tuple messages will be updated.
Fig.~\ref{topic} demonstrates that most topics extracted by ABP have almost the same top ten words as VB, GS and BP
even if ABP scans $50\%$ documents and $50\%$ topics at each iteration.

\subsection{Comparison with Online and Parallel Algorithms}

When $K=500$, $\alpha = 2/K$ and $\beta = 0.01$,
we compare ABP1 with two state-of-the-art online LDA algorithms including
OGS~\cite{Yao:09}\footnote{\url{http://mallet.cs.umass.edu/}},
OVB~\cite{Hoffman:10},\footnote{\url{http://www.cs.princeton.edu/~blei/topicmodeling.html}} and RVB~\cite{Wahabzada:11},
where we use their default parameters.
The mini-batch size is $16$ for the NIPS set,
and $1024$ for other data sets.
Based on the same test set in Table~\ref{dataset},
the algorithms OGS, OVB and RVB use the remaining $292,500$ and $8,160,000$ documents
from the original NYTIMES and PUBMED data sets~\cite{Porteous:08} for online inference.

\begin{figure*}
\centering
\includegraphics[width=0.8\linewidth]{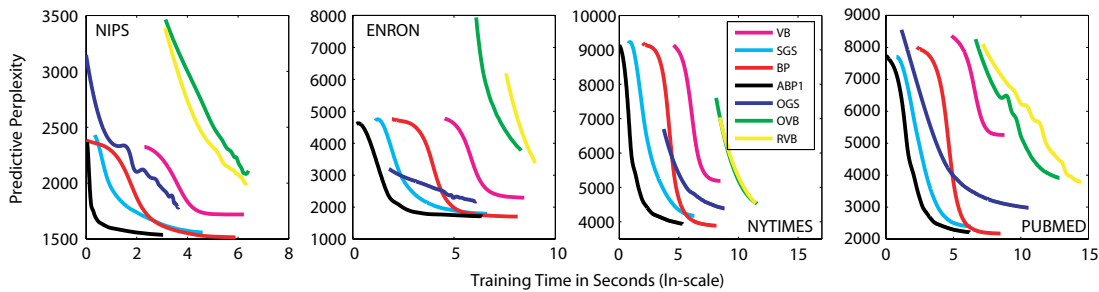}
\caption{Predictive perplexity as a function of training time in seconds ($\ln$-scale)
on NIPS, ENRON, NYTIMES and PUBMED data sets when $K=100$.}
\label{online}
\end{figure*}

Fig.~\ref{online} shows the predictive perplexity as a function of training time in seconds ($\ln$-scale) on four data sets.
Clearly,
ABP1 achieves a lower predictive perplexity by using only a fraction of training time consumed by OGS, OVB and RVB.
There are three reasons for this significant speedup.
First,
OGS is derived from SGS for each mini-batch,
which converges slower than ABP1 in Fig.~\ref{convergence}.
Second,
the computationally expensive digamma functions significantly slow down both OVB and RVB.
Finally,
OGS and OVB require loading each mini-batch from the hard disk into memory,
consuming more time than in-memory bath algorithms when the number of mini-batch is large.
RVB is slightly slower than OVB because of additional scheduling costs,
but it often achieves a relatively lower perplexity when compared with OVB.
According to~\cite{Mimno:12},
SOI is around twice faster than OVB in practice,
so ABP1 is still much faster than SOI to reach a lower predictive perplexity
when compared with the same benchmark OVB.

Fig.~\ref{online} also compares batch VB, BP and SGS with OGS, OVB and RVB.
We see that OGS, OVB and RVB perform worse than VB, BP and SGS on the NIPS and ENRON data sets,
partly because the online gradient descents introduce noises~\cite{Bottou:04} on the relatively smaller data sets.
However,
after OGS, OVB and RVB scan more training documents on the NYTIMES and PUBMED data sets,
they can achieve comparable or even lower perplexity values than SGS and VB, respectively.
These results are consistent with~\cite{Yao:09,Hoffman:10}.
Overall,
BP converges to the lowest predictive perplexity on four data sets consistent with those shown in Fig.~\ref{prediction}.
With more and more training data streams,
online algorithms may arrive at lower perplexity values than batch counterparts with significantly less memory consumptions.
However,
since memory becomes cheaper and cheaper,
RAMCloud is available and in-memory computation of batch algorithms for big data may be possible in near future~\cite{Ousterhout:09}.

\begin{table}
\centering
\caption{Average training time (seconds) per iteration.}
\begin{tabular}{|l|c|c|} \hline
Data sets &PGS ($32$ processors)      &ABP1     \\ \hline \hline
NIPS      &$0.38$    &$0.09$  \\ \hline
ENRON     &$1.69$    &$1.39$  \\ \hline
NYTIMES   &$0.68$    &$0.51$  \\ \hline
PUBMED    &$0.69$    &$0.55$  \\
\hline\end{tabular}
\label{PGS}
\end{table}

When $K=500$, $\alpha = 2/K$, $\beta = 0.01$ and $T=500$,
we also compare ABP1 with one of the state-of-the-art parallel LDA algorithms
such as PGS~\cite{Newman:09} using $32$ processors.
PGS is the parallel implementation of an approximate inference algorithm of GS
but obtains almost the same predictive perplexity values as GS in Fig.~\ref{prediction},
which indicates that the topic modeling accuracy of PGS is still lower than ABP1 except for the NIPS set.
As it would be expected,
PGS on $32$ processors should achieve $1/32$ training time of GS.
However,
PGS cannot achieve the maximum parallel efficiency due to the extensive communication and synchronization delay.
After the $32$ processors have finished scanning the distributed data sets,
they need to communicate the multinomial parameter matrix $\phi_{K \times W}$~\eqref{phiw} for synchronization.
This matrix $\phi$ increases with the number of topics $K$,
adding the burden of communication and synchronization for the large number of topics $K$.

Table~\ref{PGS} shows the average training time per iteration between ABP1 and PGS.
Obviously,
ABP1 is still much faster than PGS.
One major reason for such a speedup is that PGS searches $500$ topics at each iteration,
while ABP1 selects only $50$ topics at each iteration.
Indeed,
the large number of topics can be also handled by the parallel FGS (PFGS)~\cite{Porteous:08} and SGS (PSGS) implementations.
Although PGS can further accelerate the topic modeling speed by adding more processors,
the parallel efficiency will be further reduced due to more communication costs among more processors.
In addition,
the price-to-performance ratio of PGS is much higher than ABP1.
Overall,
ABP is a promising tool for fast topic modeling.

\section{Conclusions} \label{s6}

Fast batch LDA algorithms have attracted intensive research interests recently.
The proposed ABP algorithm accelerates training LDA significantly with a comparable topic modeling accuracy to the state-of-the-art batch LDA algorithms.
ABP scales linearly with $\lambda_dD$ and $\lambda_kK$,
and runs in sublinear time of BP.
Compared with VB~\cite{Blei:03} and GS/FGS/SGS~\cite{Griffiths:04,Porteous:08,Yao:09},
ABP achieves a much lower predictive perplexity at the faster speed.
Moreover,
ABP is even faster and more accurate than some online LDA algorithms like OGS~\cite{Yao:09} and OVB~\cite{Hoffman:10}.
Also,
ABP on a single processor is faster and more accurate than the parallel LDA algorithm like PGS~\cite{Newman:09} on $32$ processors.

In the publicly available test sets,
only $20\%$ documents and $20\%$ topics are sufficient to yield almost the same topic modeling performance as BP at each iteration,
which can be interpreted by Zipf's law or Pareto principle.
Furthermore,
ABP can choose even much smaller parameters $\lambda_d$ and $\lambda_k$ under the large $D$ and $K$.
This result implies that different documents and topics contribute differently for the topic modeling performance
in massive corpora containing a large number of topics.
For example,
at each iteration only $10\%$ documents and $10\%$ topics can provide a comparable or
much better performance than GS/FGS/SGS algorithms when $K=900$ in Fig.~\ref{prediction}.
In practice,
the subsets $\lambda_dD$ and $\lambda_kK$ can be constants
for massive data sets ($D \ge 10^6$) with a large number of topics ($K \ge 1000$).
Therefore,
ABP may use almost a fixed training time (without considering sorting cost) for probabilistic topic modeling of big data sets.
However,
as with other batch LDA algorithms,
ABP is restricted by the high memory requirement for real-world big corpora.

Due to its ease of use, high accuracy and fast speed,
ABP is a strong candidate for becoming the standard batch LDA algorithm.
Also,
ABP is a flexible EM algorithm,
which can be easily extended for other latent variable models or finite mixture models like Gaussian mixture models
and hidden Markov models~\cite{Bishop:book}.
Future work includes the relaxation of memory requirements for big corpora.
We will use the block optimization framework~\cite{Yu:10,Zeng:12b},
reading and scheduling big corpora as blocks from hard disk into memory.
With the faster speed and less memory requirement,
we may realize practical and portable topic modeling tools for big personal data sets
including documents, photos, videos on a wide range of digital platforms.

\section*{Acknowledgements} \label{s7}

This work is supported by NSFC (Grant No. 61003154, 61373092 and 61033013),
Natural Science Foundation of the Jiangsu Higher Education
Institutions of China (Grant No. 12KJA520004), Innovative Research
Team in Soochow University (Grant No. SDT2012B02) to JZ, and a GRF
grant from RGC UGC Hong Kong (GRF Project No. 9041574), a grant from
City University of Hong Kong (Project No. 7008026) to ZQL.


\begin{thebibliography}{10}
\providecommand{\url}[1]{#1}
\csname url@rmstyle\endcsname
\providecommand{\newblock}{\relax}
\providecommand{\bibinfo}[2]{#2}
\providecommand\BIBentrySTDinterwordspacing{\spaceskip=0pt\relax}
\providecommand\BIBentryALTinterwordstretchfactor{4}
\providecommand\BIBentryALTinterwordspacing{\spaceskip=\fontdimen2\font plus
\BIBentryALTinterwordstretchfactor\fontdimen3\font minus
  \fontdimen4\font\relax}
\providecommand\BIBforeignlanguage[2]{{%
\expandafter\ifx\csname l@#1\endcsname\relax
\typeout{** WARNING: IEEEtran.bst: No hyphenation pattern has been}%
\typeout{** loaded for the language `#1'. Using the pattern for}%
\typeout{** the default language instead.}%
\else
\language=\csname l@#1\endcsname
\fi
#2}}

\bibitem{Blei:12}
D.~M. Blei, ``Introduction to probabilistic topic models,''
  \emph{Communications of the ACM}, pp. 77--84, 2012.

\bibitem{Blei:03}
D.~M. Blei, A.~Y. Ng, and M.~I. Jordan, ``Latent {Dirichlet} allocation,''
  \emph{J. Mach. Learn. Res.}, vol.~3, pp. 993--1022, 2003.

\bibitem{Zeng:11}
J.~Zeng, W.~K. Cheung, and J.~Liu, ``Learning topic models by belief
  propagation,'' \emph{IEEE Trans. Pattern Anal. Mach. Intell.}, vol.~33,
  no.~5, pp. 1121--1134, 2013.

\bibitem{Bishop:book}
C.~M. Bishop, \emph{Pattern recognition and machine learning}.\hskip 1em plus
  0.5em minus 0.4em\relax Springer, 2006.

\bibitem{Asuncion:09}
A.~Asuncion, M.~Welling, P.~Smyth, and Y.~W. Teh, ``{On smoothing and inference
  for topic models},'' in \emph{UAI}, 2009, pp. 27--34.

\bibitem{Griffiths:04}
T.~L. Griffiths and M.~Steyvers, ``Finding scientific topics,'' \emph{Proc.
  Natl. Acad. Sci.}, vol. 101, pp. 5228--5235, 2004.

\bibitem{Elidan:06}
G.~Elidan, I.~McGraw, and D.~Koller, ``{Residual belief propagation: Informed
  scheduling for asynchronous message passing},'' in \emph{UAI}, 2006, pp.
  165--173.

\bibitem{Dempster:77}
A.~P. Dempster, N.~M. Laird, and D.~B. Rubin, ``{Maximum likelihood from
  incomplete data via the EM algorithm},'' \emph{Journal of the Royal
  Statistical Society, Series B}, vol.~39, pp. 1--38, 1977.

\bibitem{Thiesson:01}
B.~Thiesson, C.~Meek, and D.~Heckerman, ``Accelerating {EM} for large
  databases,'' \emph{Machine Learning}, vol.~45, pp. 279--299, 2001.

\bibitem{Zipf:book}
G.~K. Zipf, \emph{Human behavior and the principle of least effort}.\hskip 1em
  plus 0.5em minus 0.4em\relax Cambridge, Massachusetts: Addison-Wesley, 1949.

\bibitem{Porteous:08}
I.~Porteous, D.~Newman, A.~Ihler, A.~Asuncion, P.~Smyth, and M.~Welling,
  ``{Fast collapsed Gibbs sampling for latent Dirichlet allocation},'' in
  \emph{KDD}, 2008, pp. 569--577.

\bibitem{Yao:09}
L.~Yao, D.~Mimno, and A.~McCallum, ``Efficient methods for topic model
  inference on streaming document collections,'' in \emph{KDD}, 2009, pp.
  937--946.

\bibitem{Banerjee:07}
A.~Banerjee and S.~Basu, ``Topic models over text streams: A study of batch and
  online unsupervised learning,'' in \emph{SIAM Data Mining}, 2007, pp.
  437--442.

\bibitem{AlSumait:08}
L.~AlSumait, D.~Barbar{\'a}, and C.~Domeniconi, ``On-line {LDA}: Adaptive topic
  models for mining text streams with applications to topic detection and
  tracking,'' in \emph{ICDM}, 2008, pp. 3--12.

\bibitem{Canini:09}
K.~R. Canini, L.~Shi, and T.~L. Griffths, ``{Online inference of topics with
  latent Dirichlet allocation},'' in \emph{AISTATS}, 2009, pp. 65--72.

\bibitem{Hoffman:10}
M.~Hoffman, D.~Blei, and F.~Bach, ``{Online learning for latent Dirichlet
  allocation},'' in \emph{NIPS}, 2010, pp. 856--864.

\bibitem{Wahabzada:11}
M.~Wahabzada and K.~Kersting, ``Larger residuals, less work: Active document
  scheduling for latent {Dirichlet} allocation,'' in \emph{ECML/PKDD}, 2011,
  pp. 475--490.

\bibitem{Mimno:12}
D.~Mimno, M.~D. Hoffman, and D.~M. Blei, ``Sparse stochastic inference for
  latent {Dirichlet} allocation,'' in \emph{ICML}, 2012.

\bibitem{Bottou:book}
L.~Bottou, \emph{Online learning and stochastic approximations}.\hskip 1em plus
  0.5em minus 0.4em\relax Cambridge University Press, 1998.

\bibitem{Newman:09}
D.~Newman, A.~Asuncion, P.~Smyth, and M.~Welling, ``Distributed algorithms for
  topic models,'' \emph{J. Mach. Learn. Res.}, vol.~10, pp. 1801--1828, 2009.

\bibitem{Wang:09}
Y.~Wang, H.~Bai, M.~Stanton, W.~Y. Chen, and E.~Chang, ``Plda: Parallel latent
  {Dirichlet} allocation for large-scale applications,'' in \emph{Algorithmic
  Aspects in Information and Management}, 2009, pp. 301--314.

\bibitem{Yan:09}
F.~Yan, N.~Xu, and Y.~Qi, ``Parallel inference for latent {Dirichlet}
  allocation on graphics processing units,'' in \emph{NIPS}, 2009, pp.
  2134--2142.

\bibitem{Smola:10}
A.~Smola and S.~Narayanamurthy, ``An architecture for parallel topic models,''
  in \emph{PVLDB}, 2010, pp. 703--710.

\bibitem{Liu:11}
Z.~Liu, Y.~Zhang, E.~Chang, and M.~Sun, ``Plda+: Parallel latent {Dirichlet}
  allocation with data placement and pipeline processing,'' \emph{ACM
  Transactions on Intelligent Systems and Technology}, p.~26, 2011.

\bibitem{Zhai:12}
K.~Zhai, J.~Boyd-Graber, N.~Asadi, and M.~Alkhouja, ``{Mr. LDA}: A flexible
  large scale topic modeling package using variational inference in
  mapreduce,'' in \emph{WWW}, 2012.

\bibitem{Mahout:10}
A.~S. Foundation \emph{et~al.}, ``{Apache Mahout},'' 2010,
  \url{http://mloss.org/software/view/144/}.

\bibitem{Heinrich:08}
G.~Heinrich, ``Parameter estimation for text analysis,'' University of Leipzig,
  Tech. Rep., 2008.

\bibitem{Kschischang:01}
F.~R. Kschischang, B.~J. Frey, and H.-A. Loeliger, ``Factor graphs and the
  sum-product algorithm,'' \emph{IEEE Transactions on Inform. Theory}, vol.~47,
  no.~2, pp. 498--519, 2001.

\bibitem{Winn:05}
J.~Winn and C.~M. Bishop, ``Variational message passing,'' \emph{J. Mach.
  Learn. Res.}, vol.~6, pp. 661--694, 2005.

\bibitem{Asuncion:10}
A.~Asuncion, ``{Approximate Mean Field for Dirichlet-Based Models},'' in
  \emph{ICML Workshop on Topic Models}, 2010.

\bibitem{Teh:07}
Y.~W. Teh, D.~Newman, and M.~Welling, ``A collapsed variational {Bayesian}
  inference algorithm for latent {Dirichlet} allocation,'' in \emph{NIPS},
  2007, pp. 1353--1360.

\bibitem{Zeng:12a}
J.~Zeng, X.-Q. Cao, and Z.-Q. Liu, ``Residual belief propagation for topic
  modeling,'' in \emph{ADMA}, 2012, pp. 739--752.

\bibitem{Bottou:04}
L.~Bottou and Y.~L. Cun, ``Large scale online learning,'' in \emph{NIPS}, 2004,
  pp. 217--224.

\bibitem{Yu:10}
H.-F. Yu, C.-J. Hsieh, K.-W. Chang, and C.-J. Lin, ``{Large linear
  classification when data cannot fit in memory},'' in \emph{KDD}, 2010, pp.
  833--842.

\bibitem{Zeng:12b}
J.~Zeng, Z.-Q. Liu, and X.-Q. Cao, ``Memory-efficient topic modeling,'' p.
  arXiv:1206.1147 [cs.LG], 2012.

\bibitem{Yan:12}
J.~Yan, Z.-Q. Liu, Y.~Gao, and J.~Zeng, ``Communication-efficient parallel
  belief propagation for latent {Dirichlet} allocation,'' p. arXiv:1206.2190v1
  [cs.LG], 2012.

\bibitem{Zeng:13}
J.~Zeng, Z.-Q. Liu, and X.-Q. Cao, ``Online belief propagation for topic
  modeling,'' p. arXiv:1206.1147 [cs.LG], 2012.

\bibitem{Zeng:12}
J.~Zeng, ``A topic modeling toolbox using belief propagation,'' \emph{J. Mach.
  Learn. Res.}, vol.~13, pp. 2233--2236, 2012.

\bibitem{Chang:09b}
J.~Chang, J.~Boyd-Graber, S.~Gerris, C.~Wang, and D.~Blei, ``Reading tea
  leaves: How humans interpret topic models,'' in \emph{NIPS}, 2009, pp.
  288--296.

\bibitem{Ousterhout:09}
J.~Ousterhout \emph{et~al.}, ``The case for {RAMClouds}: Scalable
  high-performance storage entirely in {DRAM},'' \emph{SIGOPS Operating Systems
  Review}, vol.~43, no.~4, pp. 92--105, 2012.

\end{thebibliography}
\end{document}